\documentclass[11pt]{article}
\usepackage[utf8]{inputenc}
\usepackage[T1]{fontenc}
\usepackage{graphicx}
\usepackage{longtable}
\usepackage{wrapfig}
\usepackage{rotating}
\usepackage[normalem]{ulem}
\usepackage{amsmath}
\usepackage{amssymb}
\usepackage{capt-of}
\usepackage{hyperref}
\usepackage[T1]{fontenc}
\usepackage{lmodern}
\usepackage{csquotes}
\usepackage[font=small,labelfont=bf, justification=justified, format=plain]{caption}
\author{Jean-Marie Chauvet}
\date{\today}
\title{Memory GAPS: Would LLMs pass the Tulving Test?}
\hypersetup{
 pdfauthor={J.-M. Chauvet},
 pdftitle={Memory GAPS: Would LLMs pass the Tulving Test?},
 pdfkeywords={},
 pdfsubject={},
 pdfcreator={Emacs 28.2 (Org mode 9.6.3)}, 
 pdflang={English}}
\makeatletter
\newcommand{\citeprocitem}[2]{\hyper@linkstart{cite}{citeproc_bib_item_#1}#2\hyper@linkend}
\makeatother

\usepackage[notquote]{hanging}
\begin{document}

\maketitle

\section*{Abstract}
\label{sec:orgfe40f7b}
The Tulving Test was designed to investigate memory performance in recognition and recall tasks. Its results help assess the relevance of the ``Synergistic Ecphory Model'' of memory and similar RK paradigms in human performance. This paper starts investigating whether the more than forty-year-old framework sheds some light on LLM's acts of remembering.
\section*{Introduction}
\label{sec:orgfe90ec0}
In his groundbreaking studies of memory, Endel Tulving (1927-2023) noted that ``one of the most compelling and salient characteristics of remembering of past events is the individual's subjective awareness of remembering'' \citeprocitem{1}{[1]}. In order to include the rememberer's recollective experience into the critical constructs in the conceptualization of remembering, Tulving suggested an ``overall pretheoretical framework'', called the \emph{General Abstract Processing System} or GAPS. This paper investigates whether the GAPS also provides insights when the subject is no longer human but a Large Language Model (LLM).

Tulving championed the distinction of \emph{episodic} from \emph{semantic} memory, successfully arguing that being functionally different, they represent separate but related systems. Both are placed on the same side of the cognitive division between \emph{declarative memory} (as episodic and semantic information can be expressed through language--e.g. repairing a bicycle) on the one hand, and \emph{skills} (which can be observed only in behavior--e.g. riding a bicycle) on the other.

\subsection*{The GAPS and the Transformer}
\label{sec:org29f4574}
In Tulving's framework, a single act of remembering forms the unit of human episodic memory. Remembering is a process that begins with the witnessing or experiencing of an episode and ends with its recollective experience or with the conversion of the remembered information into some other form, or both. The GAPS specifies so called \emph{elements} of remembering and their interrelations in order to decompose this process.

The GAPS distinguishes two kinds of elements: observable events and hypothetical constructs (processes and states); and it divides elements into two categories: elements of encoding and elements of retrieval.

\begin{figure}[htbp]
\centering
\includegraphics[width=.9\linewidth]{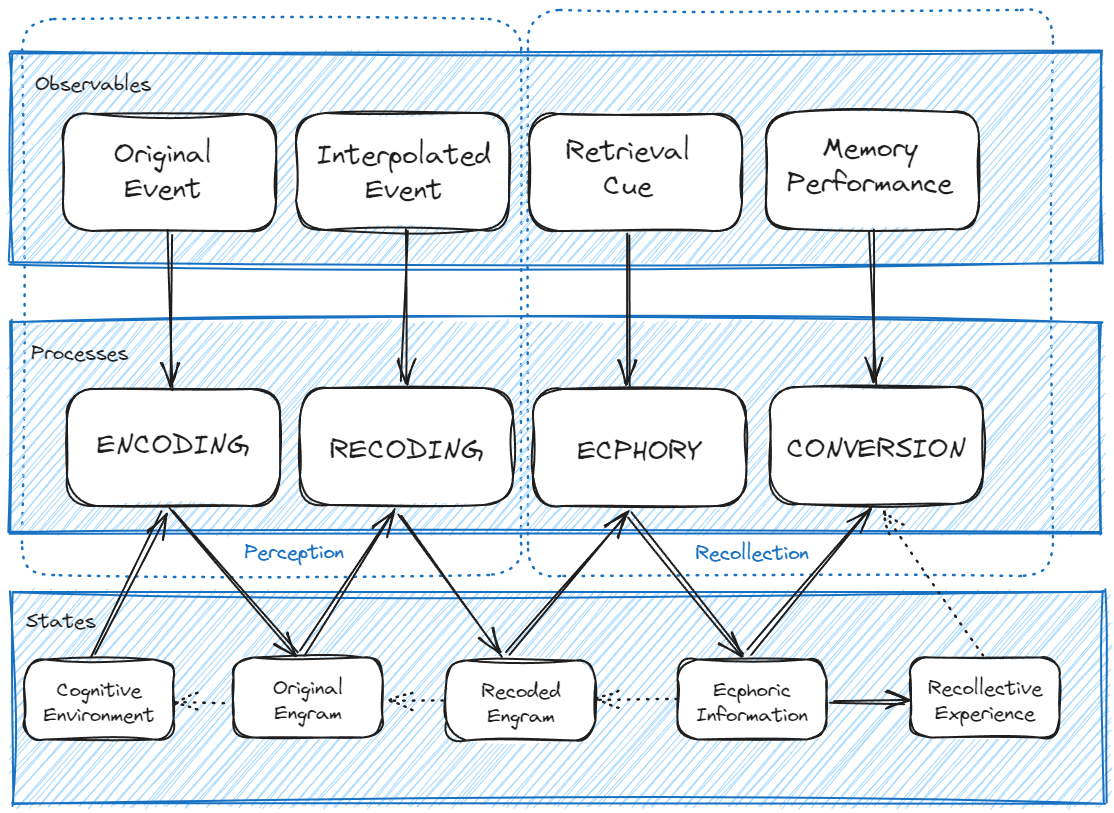}
\caption{\label{fig:orgc1390c9}\textbf{The GAPS: Elements of Episodic Memory and their Relations.} The element of encoding is a process that converts the information about an experienced event or episode (in a particular setting, at a particular time) into an \emph{engram} or memory trace. The central element of the retrieval processes \emph{ecphoric} information, a synergistic product of the engram and the retrieval cue, which calls on both episodic and semantic information. Source for figure: Ch. 7, \citeprocitem{1}{[1, 7-1, p. 135]}.}
\end{figure}

Of particular interest to this study of applicability of the GAPS framework to LLM are the possible transpositions of engram and ecphoric information into the domain of generative AI. In his seminal book, Tulving offers a very broad definition of engrams: ``the product of encoding'', ``conditions for recollection of the experienced event'', or ``differences between the state of the memory system before and after encoding''. The latter is closely related the original definitions of these terms introduced by Richard Semon (1859--1918): ``to represent the enduring changes brought about by the energetic effect of stimuli in the organism'' \citeprocitem{2}{[2]}, \citeprocitem{3}{[3]}. Note that if, in both clarifications, the nature of the changes are unknown, the term became nonetheless broadly known in psychology research through the later work of Karl Lashley (1890--1958) concluding, among other experimental results on neural mechanisms involved in learning and memory, that ``there is no demonstrable localization of memory trace'' \citeprocitem{4}{[4]}.

Similarly inspired by Semon, Tulving suggested the terms \emph{ecphory} and \emph{ecphoric information} to designate respectively the process that brings (i) the relevant information in the retrieval environment into interaction with (ii) the original or re-coded engram, and the output of this process. Such ecphoric information determines the particulars of recollective experience in the next phase of remembering: \emph{conversion}. In the GAPS model, ecphoric information is basically a task-free component of the retrieval process, it is simply used by being converted into another form in the memory performance.

The categories of encoding and retrieval in the GAPS are not without analogies with the \emph{Transformer} architecture of neural networks at the core of LLMs, which precisely articulates encoders and decoders to process vector embeddings representing words and sentences.

\begin{figure}[htbp]
\centering
\includegraphics[width=.9\linewidth]{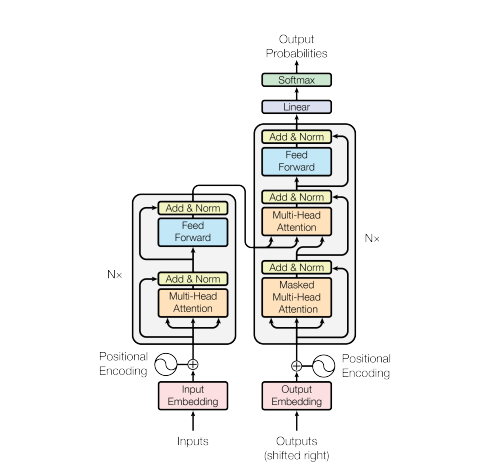}
\caption{\label{fig:org9fdc005}\textbf{The Transformer Architecture.} Based on the 2017 paper \citeprocitem{5}{[5]} attention mechanism, the Transformer architecture requires less training time than previous recurrent neural architectures. Input text is split into tokens (sometimes called \emph{n-gram}, dangerously reminiscent of Semon's engrams--see text), then converted into vectors. Through different layers, each token is contextualized with other tokens via parallel attention heads, calculating weights for each according to its importance. The Transformer Architecture elaborates on \emph{softmax-based} attention mechanism \citeprocitem{6}{[6]} and \emph{Fast Weight Controllers} \citeprocitem{7}{[7]}. Source for figure: \citeprocitem{5}{[5]}.}
\end{figure}

At this stage, from cursorily reviewing the architecture of both GAPS and Transformer--and keeping in mind that Tulving's psychological framework is only ``pre-theoretical'' and ``highly schematic'', while Transformers are actual computer implementations--the practical analogy would unfold as follows:

\begin{table}[htbp]
\caption{\label{tab:org8227773}\textbf{Hypothetical analogy between GAPS and Transformer.} Semantic memory, in Tulving's conception, would be represented by the probability distribution learned by the LLM during the pretraining phase. In Transformers it determines the particulars of the output based on the input (prompt).}
\centering
\begin{tabular}{lll}
 & GAPS & Transformer\\[0pt]
\hline
Processes & encoding & encoder\\[0pt]
 & re-coding & encoder\\[0pt]
 & ecphory & encoder\\[0pt]
 & conversion & decoder\\[0pt]
\hline
States & engram & vector embedding\\[0pt]
 & ecphoric information & output probabilities\\[0pt]
 & memory performance & output\\[0pt]
\end{tabular}
\end{table}

\subsection*{Tulving's ``direct comparison'': recognition versus recall}
\label{sec:orgecd5b4a}
In order to further investigate the analogy and its grounds, we adopt Tulving's design of ``direct comparison'' experiments to assess recognition versus recall tasks in LLMs. Recognition and recall are both processes of retrieval and both result in the rememberer's awareness of a past event. The simple episodes in the experiment are to be presentations of a list of English words to be remembered. In this simplified situation of comparing recognition and recall tasks, we consider only two independent dimensions: one has to do with the type of retrieval information, or \emph{cue}, available to the rememberer; the second refers to the conversion process in the GAPS framework. The retrieval information includes copies of the studied words and non-copy cue words. As for the conversion process: in the recognition task, the rememberer has to express whether or not the cue word was in the study list (\emph{familiarity}); in the recall task, the rememberer has to identify a word in the study list, if any, associated with the cue word (\emph{identification}), thereby expressing some other aspect of the original memorizing experience. Note that in the GAPS framework, the first dimension involves processes anterior to the construction of ecphoric information, while the second relates to post-ecphoric processes. The experimental results are therefore captured by the 2 x 2 matrix in Table \ref{tab:orgd094067}

\begin{table}[htbp]
\caption{\label{tab:orgd094067}Differences between recognition and recall tasks. Source for table: Ch. \citeprocitem{1}{[1, p. 14]}.}
\centering
\begin{tabular}{lll}
Retrieval information & Conversion & \\[0pt]
 & Familiarity & Identification\\[0pt]
\hline
Copy Cue Word & \emph{Recognition} & \emph{?}\\[0pt]
Non-Copy Cue Word & \emph{?} & \emph{Recall}\\[0pt]
\end{tabular}
\end{table}

Conventional recognition and recall tests sit in two of the four cells in the matrix. When the rememberer, however, declares a cue other than a copy cue word to be familiar it is a \emph{false positive} response from the conventional perspective although psychologists might disagree on how to think about such responses \citeprocitem{8}{[8]}. The other empty cell represents a situation where the rememberer's somewhat strange task is to repeat the cue word to confirm it is associated with the copy in the study list. \emph{False negatives} are of interest here and Tulving's interpretation was that these entailed a form of continuity between recognition and recall retrieval processes.

The direct comparison test design represents all four cells of the matrix. In a typical session the LLM is prompted to memorize a list of 48 common English words. In a group of experiments, the LLM is prompted with a cue word and asked whether the cue is included or not in the studied list; in another group, the LLM is prompted with a cue word and asked to retrieve any strongly associated word in the studied list (or none if no such word is evoked by the cue).

In each experiment 32 cue words are presented in the 32 prompts: eight of these cue words were identical with eight words in the list (\emph{copy cues}), eight were strongly associated words (\emph{non-copy associated} cues), eight were rhyming words (\emph{non copy rhymes} cues), and eight were unrelated distractors (\emph{non-copy unrelated} cues). The 32 cue words are identical for both the recognition and the recall task.

In order to introduce the distinction between immediate and delayed retrieval of the original experimental design, the experiment is run twice for each group: in the first run, memorization and retrieval are both in each individual prompt (immediate); in the second, memorization is the first prompt of a conversation (chat) with the LLM, followed by retrieval prompts which continue the conversation (delayed).

\section*{Results}
\label{sec:org710de1b}
As a reference benchmark, the results of Tulving's original experiments are presented in Table \ref{tab:org35e481b} from Ch. \citeprocitem{1}{[1, p. 14, Table 14.2]}:

\begin{table}[htbp]
\caption{\label{tab:org35e481b}\textbf{Summary of memory performance in the original direct comparison experiment.} Each proportion shown is based on 576 observations. The data for the familiarity (recognition) task show proportion of cases in which the human subjects regarded the cue word as included in the list. Hence the data for copy cues represent 'correct' responses, whereas the data from the other three types of cues represent 'false positives'. The data for the identification (recall) task indicate proportions of responses to the cue being any target word in the list.}
\centering
\begin{tabular}{l|rr|rr}
Retrieval information & Conversion &  &  & \\[0pt]
 & Familiarity &  & Identification & \\[0pt]
 & Immediate & Delayed & Immediate & Delayed\\[0pt]
\hline
Copy Cue Word & 0.78 & 0.71 & 0.69 & 0.60\\[0pt]
Non-Copy Associated & 0.15 & 0.20 & 0.54 & 0.37\\[0pt]
Non-copy Rhyme & 0.09 & 0.15 & 0.20 & 0.31\\[0pt]
Non-copy Unrelated & 0.08 & 0.18 & 0.04 & 0.02\\[0pt]
\end{tabular}
\end{table}

The memory performance of LLMs in the Tulving Test of direct comparison is presented along the same format in Table \ref{tab:orgc0ab057}.

\begin{table}[htbp]
\caption{\label{tab:orgc0ab057}\textbf{Summary of memory performance of the \texttt{mistral-7b-instruct-v0} LLM in the direct comparison experiment.} Each proportion is based on 384 observations (but see text). Interpretations of proportions are the same as above Table \ref{tab:org35e481b}.}
\centering
\begin{tabular}{l|rr|rr}
Retrieval information & Conversion &  &  & \\[0pt]
 & Familiarity &  & Identification & \\[0pt]
 & Immediate & Delayed & Immediate & Delayed\\[0pt]
\hline
Copy Cue Word & 1 & 0.46 & 0.46 & 0\\[0pt]
Non-Copy Associated & 0 & 0.47 & 0.49 & 0.40\\[0pt]
Non-copy Rhyme & 0 & 0.50 & 0.18 & 0.01\\[0pt]
Non-copy Unrelated & 0 & 0.41 & 0.08 & 0\\[0pt]
\end{tabular}
\end{table}

Within each result table, several comparisons are of interest. First the probability that copy cues were familiar was higher than the probability of identification and production of the target word in response to the copy cue, in both the human (Table \ref{tab:org35e481b}) and the  LLM (Table \ref{tab:orgc0ab057}) subject--here \texttt{mistral-7b-instruct-v0}. Second, the probability that extra-list unrelated cues were (incorrectly) recognized as members of the memorized list increased from the immediate to delayed test, in both human and LLM subjects. Remarkably and contrasting with the human subject, in the immediate recognition task the LLM never erred: no false positives for non-copy cues and 100\% familiarity for copy cues. Third, rhyme words proved in both cases more effective than unrelated distractor cues in recall. Fourth, strongly associated cues were considered members of the list with much higher probability in the immediate test, the difference being greatly reduced in the delayed test. The case of the LLM subject varies a bit, since no false positives are produced in the immediate recognition test, while they appear with similar probabilities in the delayed recognition test.

Stating the obvious when comparing the two tables: first, the LLM performs immediate recognition faultlessly, while displaying much weaker performance than human subjects on the delayed recognition: lower probability on copy cues, and significantly higher probabilities of false positives (i.e. judging non-copy cues to be included in the list). Second, in the immediate recall task the LLM memory performance is weaker than in the human subject, more so for copy cues--which fail completely--than for associate and unrelated cues--which seems paradoxical given the perfect match in the recognition task. The LLM, however, displays an intriguing pattern on the delayed identification task, comparable to human subjects when the cue is an associate word but unable to recall any word in the list when the the cued prompt is a rhyme word. The discussion section looks into the context length and so-called \emph{hallucination} phenomena as a possible cause for this last observation.

In a separate series of tests designed to investigate a different aspect of the list-memorizing episode, the recall test only was re-run with cues now being ordinals (``\emph{What is the first word in the list?}''), ranging from first to twentieth. The results are summarized in Table \ref{tab:orgd1b581a} for the immediate and delayed tests.

\begin{table}[htbp]
\caption{\label{tab:orgd1b581a}\textbf{Summary of memory performance of the \texttt{mistral-7b-instruct-v0} LLM in the direct comparison experiment with ordinal cues.} Each proportion is based on 60 observations. Interpretations of proportions are the same as above Table \ref{tab:org35e481b}.}
\centering
\begin{tabular}{lll}
Retrieval information & Conversion & \\[0pt]
 & Ordering & \\[0pt]
 & Immediate & Delayed\\[0pt]
\hline
Ordinal cue word & 0.233 & 0\\[0pt]
\end{tabular}
\end{table}

\section*{Discussion}
\label{sec:org5460900}
The LLM ``Tulving Test'' rests on the ``encoding/retrieval paradigm'' championed bu Tulving. In the paradigm, both encoding and retrieval conditions are experimentally manipulated in order to reveal specificities of each. The results presented in the previous section are part of a program of experiments to explore the relevance of these specificities, as expressed in the GAPS model above, to the current crop of LLMs in Generative AI.

In the original direct comparison experiment on human subjects, the effect of copy cues was first discussed. The finding then presented a paradox. The level of of performance with copy cues, i.e. cue words taken from the previously studied list, was generally higher in the familiarity conversion than for the identification conversion (see Table \ref{tab:org35e481b}). How can the rememberer have difficulty identifying the name of a studied item in the list, when it is the \emph{same} word which is used in the cue in the recall test, whereas it is asserted as a member of the list in the recognition test? Tulving suggests two possibilities: (i) because of differences in task requirements, the nature of the ecphoric process is different for the two groups of rememberers; and (ii) different types of conversion require different kinds, different amounts, or both of ecphoric information, namely that judgments of identification of a particular aspect of an experienced episode requires ecphoric information of ``higher quality'' than judgments that the cue word is familiar. Tulving favored the latter over the former and suggested a theory of \emph{conversion thresholds} for different kinds of memory tasks.

\begin{figure}[htbp]
\centering
\includegraphics[width=.9\linewidth]{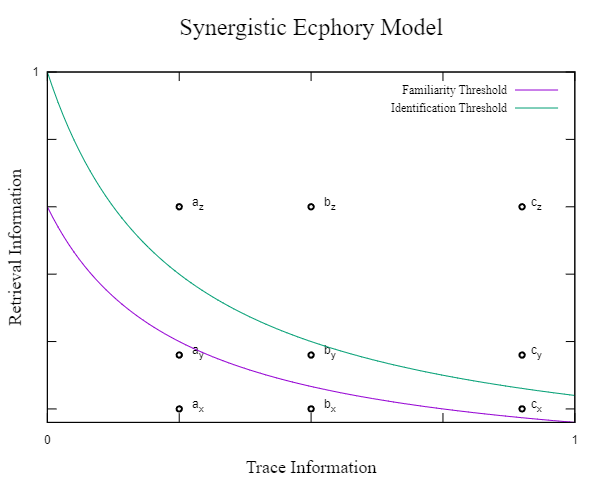}
\caption{\label{fig:org8aced19}\textbf{The Synergistic Ecphory Model (SEM)}. Source for figure: Ch.14 \citeprocitem{1}{[1, 14.3]}. Schematic diagram depicting a given episode such as the appearance of a familiar word in a particular study list. Variations of trace information, \emph{a}, \emph{b} and \emph{c} correspond to different engrams resulting from many different possible encodings of the same event, only some of which are realized on a particular occasion. Retrieval information \emph{x}, \emph{y}, \emph{z} correspond to the different potentially relevant retrieval cues (recall) that may or may not be present on a particular occasion. (We simplify here under the assumption of a single dimension for memory traces and a single dimension for retrieval information, resulting in a bi-dimensional ecphoric vector space.) The curved lines represent conversion thresholds for different memory tasks, here recognition/familiarity and recall/identification. According to the position of the point representing the synergy of retrieval and trace information relative to the threshold lines, the rememberer would pass or not the given test.}
\end{figure}

The situation with LLM rememberers is somewhat exacerbated as to the effects of copy cues. Not only the probability of recognition is higher than the same in the human test, but it is a perfect 100\%. In complete contrast, the decrease gap in performance between recognition and recall immediate tests with copy cues is much larger in the case of the LLM. And in the delayed recall test, the LLM systematically fails on copy cues.

Recent AI research on prompt engineering \citeprocitem{9}{[9]} may point to the first of Tulving's suggestions as a sensible explanation of the aggravated findings for the LLM rememberer. Different structures in the prompts, expressing recognition requests versus recall requests, may indeed entail different ecphoric processes rather than different quantity of information being required \citeprocitem{10}{[10]}. On the other hand, the conversion thresholds in the SEM are an attractive alternative as they are grounded into an ecphoric vector space, which seems cogent to the vector space embedding found in the Transformer architecture. (The explanation or the interpretation of such embedding spaces remains crucial for LLM acceptability in general \citeprocitem{11}{[11]}).

The effects of associative cues (non-copy cues) also call for discussions. Data in Table \ref{tab:org35e481b} showed a dissociation between the tasks when associative cues were used. Judgments of associative cues as included in the study list (false positives) increased from the immediate to the delayed test, whereas their effectiveness at eliciting the target word, in the recall task, decreased. What Table \ref{tab:orgc0ab057} first shows is that, even more acutely, the false positives appear only in the chat-based delayed test of recognition, whereas the LLM rememberer failed to pass any chat-based delayed test of identification. In order to quantify the correlation, Tulving introduced the measure of \emph{cue valence} \citeprocitem{12}{[12]} with respect to an aspect of an event or episode refers to the probability with which that aspect of the event can be recalled in the presence of the cue. Table \ref{tab:org35e481b} allows the quantification of the negative correlation between false positive response rate and the identification valence of associative cues. The SEM above makes also sense here, in explaining out the negative correlation by pinpointing the position in the ecphoric space relative to the conversion threshold curves. In Table \ref{tab:orgc0ab057} however, the memory performance of the delayed recall tasks appears quite contrasted: a copy cue does not recall a single word of the study list, even though the cue is itself that target word! Rhyme cue words fail almost systematically at evoking a target word in the list. Associate cue words, in contrast, trigger comparable, if slightly better, recall performance than in human subjects and unrelated are\ldots{} unrelated.

Data from Table \ref{tab:orgd1b581a} may suggest that for LLMs the role of retrieval information might be at variance with its role in human memory ecphory as posited by Tulving. In particular, examination of individual sessions with ordinal cues reveals that:
\begin{itemize}
\item In immediate tests, the first four or five answers are generally correct, i.e. the LLM properly respond to independent prompting of the first to the fifth word in the study list. Higher ordinals, however, elicit repetition of the last correct word or a random word of the list.
\item In delayed tests, LLM answers are taken consistently from another list than the study list. In one run, for instance, the LLM responses were all without repetition edible fruits, certainly picked up from some pretraining data.
\end{itemize}

Similarly, in delayed recall tests of LLM with copy cue words, the responses were mostly correct associate words, whether or not included in the study list, as if the pretrained associations took over the episodic memory traces.

Tulving contemplated that ``the rememberer's recollective experience derive its contents not only from the engram in episodic memory but also from the retrieval cue as interpreted by the semantic system and the general cognitive environment in which retrieval occurs'': the synergistic ecphory model \citeprocitem{1}{[1]}. Along these lines, the hypothesis that the joint contribution of retrieval cue and engram information might be differently balanced in LLMs and human subjects would account for the findings of Tables \ref{tab:orgc0ab057} and \ref{tab:orgd1b581a}.

The direct comparison Tulving Test when taken by LLMs result in memory performance generally comparable to human memory performance with significant extreme differentiation in the immediate recognition test, and in the delayed recall test, in particular when prompted with copy cues or ordinal cues. (Rhyme cue words are also much less efficient in LLM recall.) Within the GAPS framework and associated Synergistic Ecphory Model, these data suggest that semantic memory information, built by pretraining a LLM, and imported by cues into the retrieval information outweighs the engram information of episodic memory.

In a famous paper of 1980, William K . Estes (1919-2011), asked ``Is Human Memory Obsolete?'' \citeprocitem{13}{[13]} noting that: ``We evidently can conclude with some confidence, then, that a person's memory for elements of a sequence of items such as letters, digits, or words is best represented by uncertainty gradients portraying the way information about the remembered position of each item is distributed over an interval of time, rather than by a series of boxes or slots containing items of information.'' Although Estes insisted there on the difference between probability distributions for human memory traces versus on/off deterministic slot structures of raw computer memory, today's Transformer architecture--which execution indeed uses the same on/off computer memory--revolves also around probability distributions and vector spaces, as in Estes's human memory model. The Tulving Test helps qualifying this analogy in memory performance, hinting at a different balance in usage of semantic information and cue/retrieval information in delayed recall of episodic memories in LLMs and human subjects. Further work is required for a quantitative assessment of the variant ecphory processes in human subjects and LLMs, more specifically on a possible analytical definition of Tulving's cue valence based not on an estimate of observed memory performance but directly on the general probability distributions learned by LLMs in pretraining.

\section*{Methods}
\label{sec:org92b3ce7}
We transpose the direct comparison experiment, between recognition and recall, described in \citeprocitem{1}{[1, 14]} to LLM subjects.

Individual experiments are programmed as Python scripts interacting with LLMs through the LLM CLI utility and library \citeprocitem{14}{[14]} (Python 3.11.8 on Windows 10). Results presented and discussed in this paper were obtained with \texttt{mistral-7b-instruct-v0} \citeprocitem{15}{[15]}. (Results with smaller models, e.g. \texttt{orca-mini-3b} \citeprocitem{16}{[16]}, were not reliable enough.)

48 simple English words were selected manually to constitute the study list of to-be-remembered words. Firstly, 48 associate cue words were selected from three sources: (i) prompting the LLM for one strongly associated word to each of the 48 to-be-remembered words, (ii) synonyms of each of the 48 words, and (iii) antonyms of each of the 48 words. Antonyms and synonyms were obtained using the Natural Language Toolkit \citeprocitem{17}{[17]}. Secondly, 48 rhyme cue words were obtained using the CMU Pronouncing Directory \citeprocitem{18}{[18]}. Finally, 16 unrelated English words were picked up manually to act as distractors. The 48-row by 3-column table of target word, associate cue word, rhyme cue words together with the list of 16 distractors is the product of these initial preparation scripts.

Each session is made of two tests, one on the recognition task (familiarity), the other on the recall task (identification). Each test lists 32 cue words submitted to the LLM for remembering either (i) if the cue word is included in the study list, for recognition, or (ii) a word in the study list evoked by the cue word, or ``none'' (recall). The 32 cue words are grouped into 8 copy cues, 8 associate cues, 8 rhyme cues and 8 unrelated cues. Both the order of the 32 cues and the selection of cue types are randomized before running each session.

The recognition and recall 32-word tests are run twice to differentiate immediate from delayed performance. In immediate tests, each individual prompt to the LLM contains the list of 48 words to be remembered before the cue word. In delayed test, each test is a chat beginning with the first instruction to memorize the list of 48 words, preceding a series of individual prompts for each cue word, all within the same chat.

Each response of the LLM is analyzed and two counts are updated for the presence of the target word in the response, and for the presence of any word in the study list. (Note that the second count deliberately includes false positives in the recognition task with non-copy cues.)

\section*{References}
\label{sec:orgfc03513}

\noindent \hypertarget{citeproc_bib_item_1}{[1] E. Tulving, \textit{Elements of Episodic Memory}. Oxford University Press, 1983.}

\noindent \hypertarget{citeproc_bib_item_2}{[2] D. L. Schacter, J. E. Eich, and E. Tulving, “Richard Semon’s Theory of Memory,” \textit{Journal of Verbal Learning and Verbal Behavior}, vol. 17, no. 6, pp. 721–743, 1978, doi: \url{https://doi.org/https://doi.org/10.1016/S0022-5371(78)90443-7}.}

\noindent \hypertarget{citeproc_bib_item_3}{[3] R. W. Semon, \textit{Die Mneme als erhaltendes Prinzip im Wechsel des organischen Geschehens}. Engelmann, 1920. doi: \href{https://doi.org/10.5962/bhl.title.10234}{10.5962/bhl.title.10234}.}

\noindent \hypertarget{citeproc_bib_item_4}{[4] K. S. Lashley, “In Search of the Engram,” 1950.}

\noindent \hypertarget{citeproc_bib_item_5}{[5] A. Vaswani \textit{et al.}, “Attention is all you need,” in \textit{Proceedings of the 31st International Conference on Neural Information Processing Systems}, in Nips’17. Long Beach, California, USA: Curran Associates Inc., 2017, pp. 6000–6010.}

\noindent \hypertarget{citeproc_bib_item_6}{[6] D. Bahdanau, K. Cho, and Y. Bengio, “Neural Machine Translation by Jointly Learning to Align and Translate.”}

\noindent \hypertarget{citeproc_bib_item_7}{[7] J. S. Imanol Schlag Kazuki Irie, “Linear Transformers Are Secretly Fast Weight Programmers.”}

\noindent \hypertarget{citeproc_bib_item_8}{[8] M. K. Moshe Anisfeld, “Association, Synonymity, and Directionality in False Recognition,” \textit{Journal of Experimental Psychology}, vol. 77, no. 2, p. 171, 1968, doi: \href{https://doi.org/10.1037/h0025782}{10.1037/h0025782}.}

\noindent \hypertarget{citeproc_bib_item_9}{[9] J. Wei \textit{et al.}, “Chain-of-thought prompting elicits reasoning in large language models.” 2023.}

\noindent \hypertarget{citeproc_bib_item_10}{[10] J. Li and J. Li, “Memory, Consciousness and Large Language Model.”}

\noindent \hypertarget{citeproc_bib_item_11}{[11] G. Tennenholtz \textit{et al.}, “Demystifying Embedding Spaces using Large Language Models.”}

\noindent \hypertarget{citeproc_bib_item_12}{[12] E. Tulving and M. J. Watkins, “Structure Of Memory Traces,” \textit{Psychological review}, vol. 82, no. 4, pp. 261–275, Jul. 1975, doi: \href{https://doi.org/10.1037/h0076782}{10.1037/h0076782}.}

\noindent \hypertarget{citeproc_bib_item_13}{[13] W. K. Estes, “Is human memory obsolete?,” \textit{Am sci}, vol. 68, no. 1, pp. 62–69, Jan. 1980.}

\noindent \hypertarget{citeproc_bib_item_14}{[14] S. Willison, “LLM.” https://llm.datasette.io/en/stable/index.html, 2023.}

\noindent \hypertarget{citeproc_bib_item_15}{[15] A. Q. Jiang \textit{et al.}, “Mistral 7B.” https://huggingface.co/mistralai/Mistral-7B-v0.1}

\noindent \hypertarget{citeproc_bib_item_16}{[16] P. Mathur, “An explain tuned OpenLLaMA-3b model on custom wizardlm, alpaca, and dolly datasets,” \textit{Github repository, huggingface repository}. https://github.com/pankajarm/wizardlm\_alpaca\_dolly\_orca\_open\_llama\_3b, https://https://huggingface.co/psmathur/wizardlm\_alpaca\_dolly\_orca\_open\_llama\_3b; GitHub, HuggingFace, 2023.}

\noindent \hypertarget{citeproc_bib_item_17}{[17] S. Bird, E. Klein, and E. Loper, \textit{Natural language processing with Python: analyzing text with the Natural Language Toolkit}. O’Reilly Media, Inc., 2009.}

\noindent \hypertarget{citeproc_bib_item_18}{[18] Carnegie Mellon Speech Group, “The CMU Pronouncing Dictionary.” }\bigskip

\section*{Author information}
\label{sec:orge5de7d4}
Jean-Marie Chauvet is a co-founder of Neuron Data and served as its CTO (1985-2000). He no longer maintains any affiliation.

J.-M. C. performed all analyses and wrote the manuscript as an independent researcher.

\section*{Ethics declarations}
\label{sec:org1e293fd}
The author declare no competing interests.

\section*{Electronic supplementary material}
\label{sec:org4c0d78b}
\subsection*{Data Availability}
\label{sec:org103c309}
Results of the Tulving Tests analysed in the paper are publicly available in the repository:
\url{https://github.com/CRTandKDU/TulvingTest/tree/main/tulving/output}

For each complete session, immediate and delayed recognition and recall tests are kept in separate CSV files.

\subsection*{Code Availability}
\label{sec:org69890a3}
Python scripts for the Tulving Test and tabulation of their results are publicly available in the repository:
\url{https://github.com/CRTandKDU/TulvingTest/tree/main/tulving}
\end{document}